\documentclass[10pt,twocolumn,letterpaper]{article}

\usepackage{wacv}              

\usepackage{graphicx}
\usepackage{amsmath}
\usepackage{amssymb}
\usepackage{booktabs}
\usepackage{amsfonts}
\usepackage{mathrsfs}
\usepackage{tabularx}
\usepackage{multirow}
\usepackage{ulem}

\usepackage[pagebackref,breaklinks,colorlinks]{hyperref}
\usepackage{indentfirst}

\usepackage[capitalize]{cleveref}
\crefname{section}{Sec.}{Secs.}
\Crefname{section}{Section}{Sections}
\Crefname{table}{Table}{Tables}
\crefname{table}{Tab.}{Tabs.}

\def\wacvPaperID{1126} 
\def\confName{WACV}
\def\confYear{2024}

\begin{document}


\title{HDMNet: A Hierarchical Matching Network with Double Attention for Large-scale Outdoor LiDAR Point Cloud Registration}
\author{Weiyi Xue,~Fan Lu,~Guang Chen\thanks{Corresponding author: guangchen@tongji.edu.cn}\\
Tongji University\\
{\tt\small \{xwy,lufan,guangchen\}@tongji.edu.cn}
}

\maketitle    

\maketitle
\begin{abstract}
   Outdoor LiDAR point clouds are typically large-scale and complexly distributed. To achieve efficient and accurate registration, emphasizing the similarity among local regions and prioritizing global local-to-local matching is of utmost importance, subsequent to which accuracy can be enhanced through cost-effective fine registration. In this paper, a novel hierarchical neural network with double attention named HDMNet is proposed for large-scale outdoor LiDAR point cloud registration. Specifically, A novel feature consistency enhanced double-soft matching network is introduced to achieve two-stage matching with high flexibility while enlarging the receptive field with high efficiency in a patch-to-patch manner, which significantly improves the registration performance. Moreover, in order to further utilize the sparse matching information from deeper layer, we develop a novel trainable embedding mask to incorporate the confidence scores of correspondences obtained from pose estimation of deeper layer, eliminating additional computations. The high-confidence keypoints in the sparser point cloud of the deeper layer correspond to a high-confidence spatial neighborhood region in shallower layer, which will receive more attention, while the features of non-key regions will be masked. Extensive experiments are conducted on two large-scale outdoor LiDAR point cloud datasets to demonstrate the high accuracy and efficiency of the proposed HDMNet.
\end{abstract}
\section{Introduction}
\label{1}
\begin{figure}[t]
    \centering
    \includegraphics[width=0.85\linewidth]{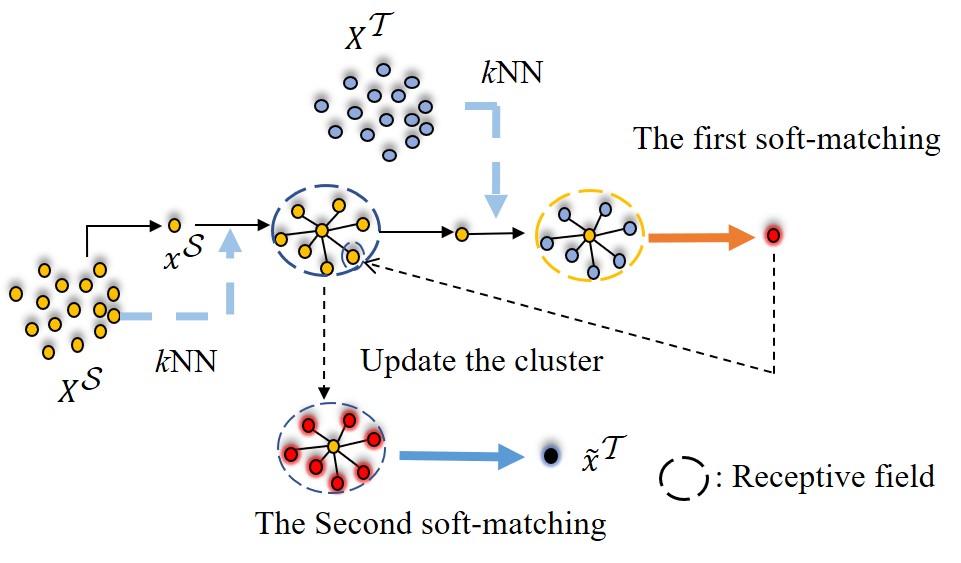}
    \caption{Illustration of double-soft matching strategy. Given the source point cloud $X^\mathcal{S}$ and the target point cloud $X^\mathcal{T}$, we perform two rounds of soft-matching. The first round of soft-matching is conducted between the source and target point clouds to update the points in the source point cloud. Subsequently, the second round of soft-matching is performed within the updated source point cloud.}
    \label{fig1}
\end{figure}
Point cloud registration is a fundamental task in 3D computer vision, aiming to estimate the optimal rigid transformation for aligning two point clouds. This task shares similarities with tasks such as LIDAR odometry\cite{elhousni2020survey}, and has been utilized in diverse practical applications such as intelligent robotics\cite{pomerleau2015review}, autonomous driving\cite{nagy2018real}. \\
\hspace*{1em}In certain registration algorithms, the performance is hindered due to the characteristics of outdoor Lidar point clouds, including their heightened sparsity, expanded spatial extent, and intricate distribution. The Iterative Closest Point(ICP)\cite{besl1992method} and its variants\cite{yang2015go,rosen2019se,serafin2015nicp} are widely recognized as the most prominent approaches. However, ICP are heavily relies on the initial alignment and easily converge to a local minimum. Most of learning-based methods primarily concentrate on object-level\cite{wang2019deep,aoki2019pointnetlk,wang2019prnet,li2020iterative} or indoor point clouds\cite{choy2020deep,huang2020feature,gojcic2020learning} and rely on assumptions regarding point cloud distributions. Recently, HRegNet\cite{lu2021hregnet} have demonstrated remarkable efficacy in addressing the challenges posed by sparse features. However, considering the presence of errors in keypoint descriptors, the utilization of point-to-point\cite{li2020iterative} and point-to-patch\cite{li2019usip,lu2020rskdd,lu2021hregnet} matching strategies can introduce erroneous correspondences. As a result, existing approaches either lack reliability or are time-consuming when applied to outdoor LiDAR point cloud registration.\\
\begin{figure*}[!t]
    \centering
    \includegraphics[width=0.98\textwidth]{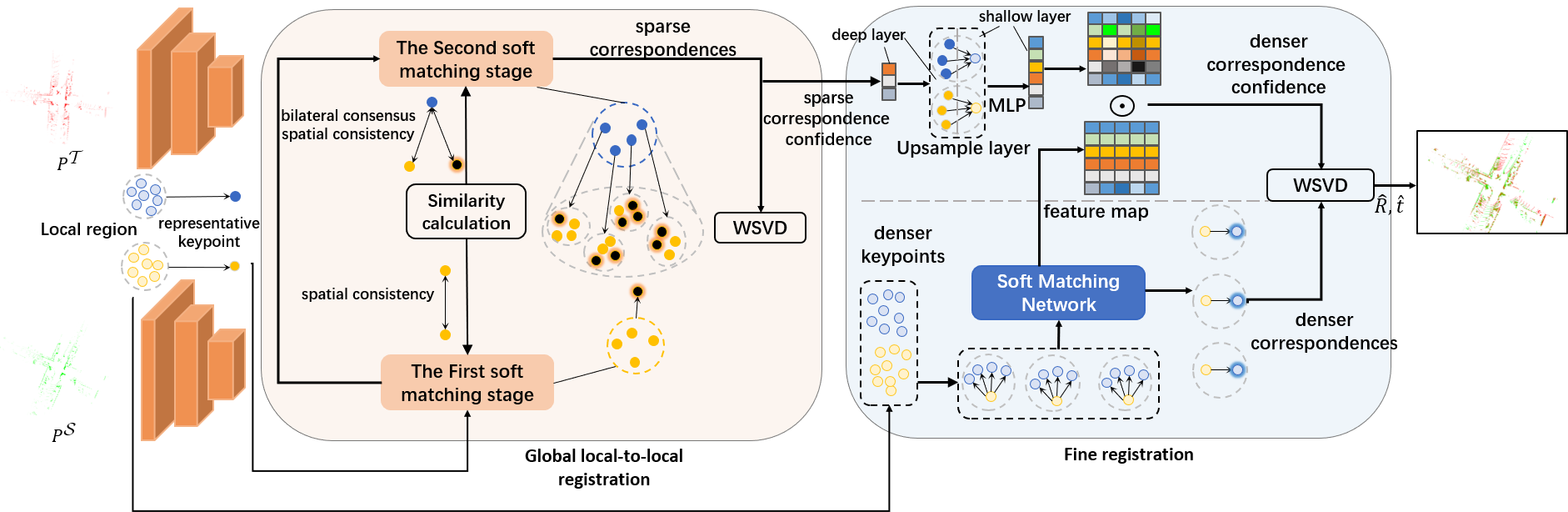}
    \caption{Network architecture of the proposed HDMNet. HDMNet consists of two main components: global local-to-local and fine registration. We use a feature extractor to hierarchically abstract the local regions into keypoints and descriptors, leveraging the reliable features at the deeper layer along with precise position information at the shallower layer. We employ the feature consistency enhanced double-soft matching module for global registration while simultaneously outputting the confidence scores of the correspondences. After that, we utilize the confidence scores to generate an initial mask for local regions, enhancing the focus on key regions and establish a denser set of correspondences to perform fine registration.}
    \label{fig2}
\end{figure*}
To address the aforementioned challenges, we propose HDMNet for large-scale outdoor point cloud registration. HDMNet performs registration hierarchically, leveraging the generation of virtual keypoints. In global local-to-local registration process of the deepest layer, considering the potential errors in descriptors that may result in a number of mismatches, we employ a double-soft matching strategy, as depicted in \cref{fig1}. This strategy involves two consecutive matching stages, wherein the points from the target point cloud participate in both matching steps. To incorporate them into the registration pipeline, we introduce a learning-based correspondence network with a feature consistency enhanced double-soft matching module, achieving two-stage matching with high flexibility while enlarging the receptive field. Furthermore, in the global registration, the sparsity of keypoints enables us to employ more robust strategies while maintaining high efficiency, thus we incorporate  consistency features and sparse-to-denser matching strategy into the double-soft matching network, where sparse-to-denser matching refers to matching the sparse source point clouds with our updated, denser target point clouds. The entire double-soft matching network significantly improves the registration accuracy due to the patch-to-patch approach to search for correspondences among super points.\\
\hspace*{1em}The overall network architecture is illustrated in \cref{fig2}. During the estimation of pose transformation in local-to-local registration, we simultaneously learn a set of confidence scores for correspondences, we can infer that each keypoint in sparser point clouds of the deeper layer correspond to a spatial neighborhood region in denser point cloud of shallower layer. Therefore, these scores, which represent the confidence of the local region of shallower layer, are reused in fine registration, thereby avoiding repetitive calculations and enhancing overall efficiency. Extensive experiments are conducted on two large-scale outdoor LiDAR point cloud datasets, namely the KITTI odometry dataset\cite{geiger2012we} and the NuScenes dataset\cite{caesar2020nuscenes}. The result demonstrate that HDMNet surpasses existing approaches in accuracy while maintaining high efficiency, presenting remarkable improvements.\\


\hspace*{1em}Overall, our contributions are as follows:
\begin{itemize}
  \item Our HDMNet for large-scale outdoor point cloud registration achieves state-of-the-art performance with high computational efficiency. 
  \item We design feature consistency enhanced double-soft matching network, achieving two-stage matching with high flexibility while enlarging the receptive field in a patch-to-patch manner, significantly improving the performance.
  \item We design a mask prediction module to prioritize crucial regions with higher correspondence confidence, effectively leveraging the confidence from upper-layer to enhance accuracy while avoiding duplicated computations.
\end{itemize}
\section{Related Works}
\label{2}
\subsection{Conventional point cloud registration}
The Iterative Closest Point (ICP)\cite{besl1992method} is widely recognized as the most prominent method for point cloud registration. However, its heavy reliance on the initial estimate and susceptibility to local minima have motivated the development of several variants\cite{yang2015go,rosen2019se,serafin2015nicp}. FGR and TEASER\cite{zhou2016fast,yang2020teaser} are tolerant to outliers from robust cost functions. Some approaches focus on feature extraction from point clouds\cite{flint2007thrift,rusu2009fast,sipiran2011harris,tombari2010unique,johnson1999using}. For instance, Fast Point Feature Histogram (FPFH)\cite{rusu2009fast} constructs an oriented histogram based on pairwise geometric properties. \cite{guo2016comprehensive} presents a review of handcrafted features in 3D point clouds. Subsequently, Random sample Consensus (RANSAC)\cite{fischler1981random} and its variants \cite{torr2000mlesac,barath2019magsac,barath2020magsac++} are commonly employed to remove outliers from the initial correspondences.
\subsection{Learning-based point cloud registration}
\noindent \textbf{End-to-end registration.} PointNetLK\cite{aoki2019pointnetlk} integrates the Lucas \& Kanade algorithm\cite{lucas1981iterative} with PointNet\cite{qi2017pointnet} to perform registration. In Feature-metric registration (FMR)\cite{huang2020feature}, the alignment of two point clouds is achieved by minimizing the error of global feature projection. A transformer network was utilized in Deep Closest Point\cite{wang2019deep} to estimate soft correspondences while introduces significant computational overhead, leading to low efficiency. IDAM\cite{li2020iterative} introduces an iterative distance-aware similarity matrix convolution module and proposes a learnable point cloud downsampling method by utilizing the hard and hybrid point elimination, which evaluate and filter each point based on the scores.\\
\\
\textbf{Feature matching-based registration and deep point correlation.} Feature matching methods commonly use precomputed features to establish point correspondences. Learning-based outlier rejection modules are then employed for correspondence filtering\cite{choy2020deep,bai2021pointdsc,lu2021hregnet,pais20203dregnet,lu2019deepvcp,gojcic2020learning}. CoFiNet\cite{yu2021cofinet} learns feature descriptors in a coarse-to-fine manner. BUFFER\cite{Ao_2023_CVPR} takes advantage of both point-wise and patch-wise techniques. GeoTransformer\cite{qin2023geotransformer} learns geometric feature for robust superpoint matching. Deep Global Registration (DGR)\cite{choy2020deep} achieves state-of-the-art performance in indoor point cloud registration by introducing Fully Convolutional Geometric Features (FCGF)\cite{choy2019fully}. DeepVCP\cite{lu2019deepvcp} leverages virtual points to establish correspondences. Furthermore, Usip\cite{li2019usip} and Rskdd-Net\cite{lu2020rskdd} employ self-supervised learning methods to generate virtual keypoints and extract descriptor features. Inspired by attentive cost volume\cite{wang2021hierarchical,wang2021pwclo,wang2022efficient}, our objective is to find point correspondences in our registration task. FlowNet3D\cite{liu2019flownet3d} introduces an embedding layer that learns point correlations in consecutive frames. Wu \textit{et al}.~\cite{wu2020pointpwc} propose a cost volume method for point clouds, incorporating individual point motion patterns. Wang \textit{et al}.~further develop the attentive cost volume method, applying it to end-to-end odometry\cite{wang2021pwclo,wang2022efficient}, which share similarities with registration. Recently, HRegNet\cite{lu2021hregnet} achieving superior accuracy and efficiency compared to previous methods. However, Erroneous correspondences can be introduced due to the presence of errors in keypoint descriptors and the utilization of point-to-patch matching strategies.
\section{Methodology}
\label{sec:method}
The input of HDMNet are source and target point clouds which are standardized to the same number of points. HDMNet makes a prediction of the optimal rotation matrix to align the source point clouds with the target point clouds. The siamese feature pyramid, which will be described in \cref{3.1}, is employed to encode the two point clouds and generate virtual keypoints along with their descriptors. Subsequently, a feature consistency enhanced double-soft matching network is introduced in \cref{3.2} to achieve global local-to-local alignment via two-stage matching in descriptor space. After that, we incorporate the confidence scores for correspondences obtained from local-to-local registration into the fine registration of the upper layer , which will be explained in \cref{3.3}. After two layers of fine registration, the final estimation $\mathbf{\hat{R}}$ and $\mathbf{\hat{t}}$ are obtained.
\subsection{Siamese Point Feature Pyramid}
\label{3.1}
We hierarchically utilize 3-layers of feature extraction, where each layer takes keypoints $X_{l-1}$ (or raw point clouds), uncertainties $\tilde{\Sigma}_{l-1}$, and descriptors $\tilde{D}_{l-1}$ as inputs , and outputs $\tilde{X}_l$, $\tilde{D}_l$, $\tilde{\Sigma}_l$. Each level's feature extraction consists of two parts: a keypoint generation module and a descriptor generation module. The detailed network structure can be referred to \cite{lu2020rskdd}. Additionally, for each level's input, we select a set of candidate keypoints using the Weighted Farthest Point Sampling (WFPS)\cite{zhou2020da4ad}, which introduces uncertainties for sampling to achieve more reliable keypoint selection. Detailed derivation of this method can be found in \cite{lu2021hregnet}.
\subsection{Global local-to-local Matching}
\label{3.2}
Due to the downsampling method employed, the keypoints in the deepest layer, which are generated by hierarchically aggregating points from local regions in the shallower layers, can be regarded as representations of local regions in the original point cloud and exhibit lower uncertainty and higher reliability. The objective of this module is to achieve local-to-local registration using the sparser keypoints. The pivotal challenge lies in establishing the correspondence between points of two point clouds. However, using a single soft matching strategy to generate matching points by \textit{k}NN search in a point-to-patch manner fails to consider sufficient information from the source point cloud and not suitable when there are limited overlaps between the two point clouds. To address these issues, we employ a learning-based correspondence network with a feature consistency enhanced double-soft matching model in the local-to-local matching process in layer 3.
\subsubsection{Double-soft matching network}
\label{3.2.1}
\begin{figure}[t]
    \centering
    \includegraphics[width=1\linewidth]{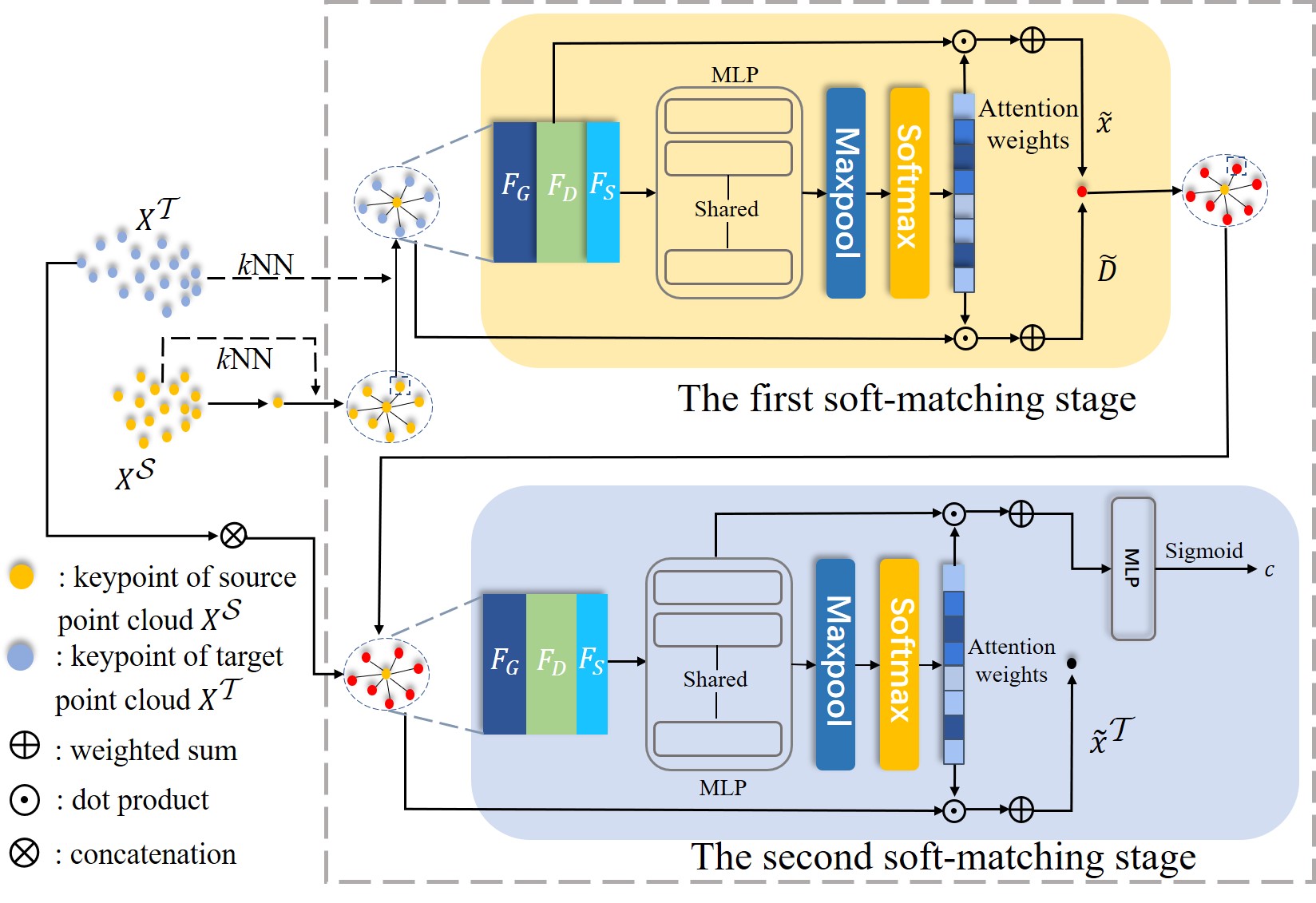}
    \caption{The detailed calculation diagram for our feature consistency enhanced double-soft matching network.The detailed process description is in \cref{3.2} }
    \label{fig3}
\end{figure}

\noindent\textbf{Double-soft matching in local-to-local registration.} To achieve the double-soft matching strategy in local-to-local registration, as illustrated in \cref{fig3}, we designed a network consisting of two soft matching modules with identical structures. The double-soft matching network takes keypoints $X^\mathcal{S}=\{x_1,x_2,...,x_i\}_{i=1}^N$, $X^\mathcal{T}=\{y_1,y_2,...,y_i\}_{i=1}^N$, descriptors $D^\mathcal{S},D^\mathcal{T}\in\mathbb R^{N\times C}$, and uncertainties $\Sigma^\mathcal{S},\Sigma^\mathcal{T}\in\mathbb R^{N\times 1}$ as inputs, we omit the subscripts $l=3$ indicating the layer number in this section. Firstly, for each $x_i$ in $X^\mathcal{S}$, we perform \textit{k}NN search in the source point cloud $X^\mathcal{S}$ to find the $k_1$ nearest neighbor points $\{x_i^1,x_i^2,...,x_i^{k_1}\}$. Then, for each neighboring point $x_i^j$, we search for $k_2$ nearest neighbor points in the target point cloud, forming $k_1$ point clusters $\{X_i^1,X_i^2,...,X_i^{k_1}\}$. Each cluster $X_i^j=\{x_i^{j_1},x_i^{j_2}, ...,x_i^{j_{k_2}}\}$ contains $k_2$ points, represents the cluster formed by the $j$-th nearest neighbor points of the $i$-th point in the source point cloud. In summary, each point $x_i$ in the source point cloud corresponds to $k_1$ patches of the target point cloud. All the aforementioned \textit{k}NN searches are conducted in the descriptor space.\\
\hspace*{1pc}The purpose of the first soft matching module is to focus on each cluster $X_i^j$ and aggregate it into a single point to update the target point cloud. This single-soft matching network is inspired by \cite{lu2020rskdd,lu2021hregnet}, where the features of points within each cluster include similarity features $F_S$, geometric features $F_G$ and descriptor features $F_D$. $F_G$, $F_D$ are described in detail in \cite{lu2021hregnet}. In our approach, $F_S$ differs from HRegNet and represents the similarity features between neighboring and center keypoints, incorporating feature consistency similarity and bilateral consensus. Our similarity calculation is simpler and differs between the two soft matching stages, which will be explained in detail in \cref{3.2.2}. We utilize a 3-layer Shared MLP to apply a nonlinear transformation to the Cluster features $X_i^j$. Subsequent maxpool and the softmax function are used to predict weights $w_i^j=\{w_i^{j_1},w_i^{j_2},...,w_i^{j_{k_1}}\}$ for all candidate keypoint within $X_i^j$. Then we use $w_i^j$ to compute weighted sum of the points within the cluster. Updated coordinates and descriptors as output is denoted as $\tilde{X}=\{\tilde{x}_1,\tilde{x}_2,...\tilde{x}_{k_1}\}$ and $\tilde{D}$, respectively.\\
\noindent\hspace*{1pc}In the second soft matching module, we focus on the $k_1$ points of $\tilde{X}$ derived from the $k_1$ clusters in the first soft matching module. Each individual point in $\tilde{X}$ represents a patch within the target point cloud, collectively forming a new cluster. While maintaining a soft matching strategy, the structure of the second soft matching network exhibits similarities to the first module but with some differences: Firstly, the computation of $F_S$ differs. Additionally, by passing the cluster features through a 3-layer Shared-MLP, we obtain a feature map $\bar{F}=\{f_1,...,f_{K_2}\}$, which will be fed into a new MLP with a sigmoid function to predict a confidence score $\tilde{c}$ for the correspondence. Finally, the calculation of the optimal transformation $\mathbf{R^*}$ and $\mathbf{t^*}$ can be achieved by solving \cref{eq:1} in a closed-form using weighted kabsch algorithm\cite{kabsch1976solution} based on the corresponding keypoints and confidence scores.
\begin{equation}\label{eq:1}
\mathbf{R^*},\mathbf{t^*} = \operatorname*{arg\,min}_{\mathbf{R}, \mathbf{t}} \sum_{i=1}^{n} \tilde{c}_i\lVert \tilde{x}_i^\mathcal{T} - (\mathbf{R}x_i^\mathcal{S} + \mathbf{t}) \rVert^2
\end{equation}
where $\lVert \cdot \rVert_2$ represents $L_2$ norm, $x_i^S$ and $\tilde{x}_i^T$ are corresponding points, $\tilde{c}_i$ represents the correspondence confidence. This double soft matching mode adopts a patch-to-patch approach, focusing on $k_1$ keypoints from the source point cloud and $k_1k_2$ keypoints from the target point cloud. In comparison to utilizing a single soft matching, it offers enhanced flexibility. Each point in the target point cloud directly participates in two soft matching processes. A higher weight in the first soft matching process indicates its ability to better represent a local region within the target point cloud. Similarly, a higher weight during the second matching process suggests a greater likelihood of this local region matching the point $x_i$ from the source point cloud $X^\mathcal{S}$.\\
\textbf{Sparse-to-Denser matching strategy.} To efficiently implement the Double-soft matching strategy described above and avoid redundant computations caused by the possibility of a point in the target point cloud being a neighbor to multiple points in the source point cloud, in our practical implementation, we reverse the order of \textit{k}NN searches. We firstly for each point in the source point cloud perform \textit{k}NN search in the target point cloud to form a cluster, aggregate the cluster into a single point to update the target point cloud. Subsequently, we conduct a same soft matching process between the source point cloud and the updated target point cloud. This computation method enables the implementation of additional sparse-to-denser matching strategies. Specifically, considering that the target point cloud was updated in the first soft matching model and its structure was also altered, resulting in the loss of some initial information of the target point cloud. Drawing inspiration from ResNet\cite{he2016deep}, a simple strategy to avoid the aforementioned issue is to concatenate the updated target and original target point cloud before the second round of soft matching. This approach ensures an increase in the density of the target point cloud while maintaining accuracy compared to using a single round of soft matching alone. Consequently, the target point cloud is composed of both the pre-update point clouds and the updated point clouds in the second round of soft matching, leading to sparse-to-denser matching(sparse source point clouds match denser target point clouds).
\begin{figure}[t]
    \centering
    \includegraphics[width=\linewidth]{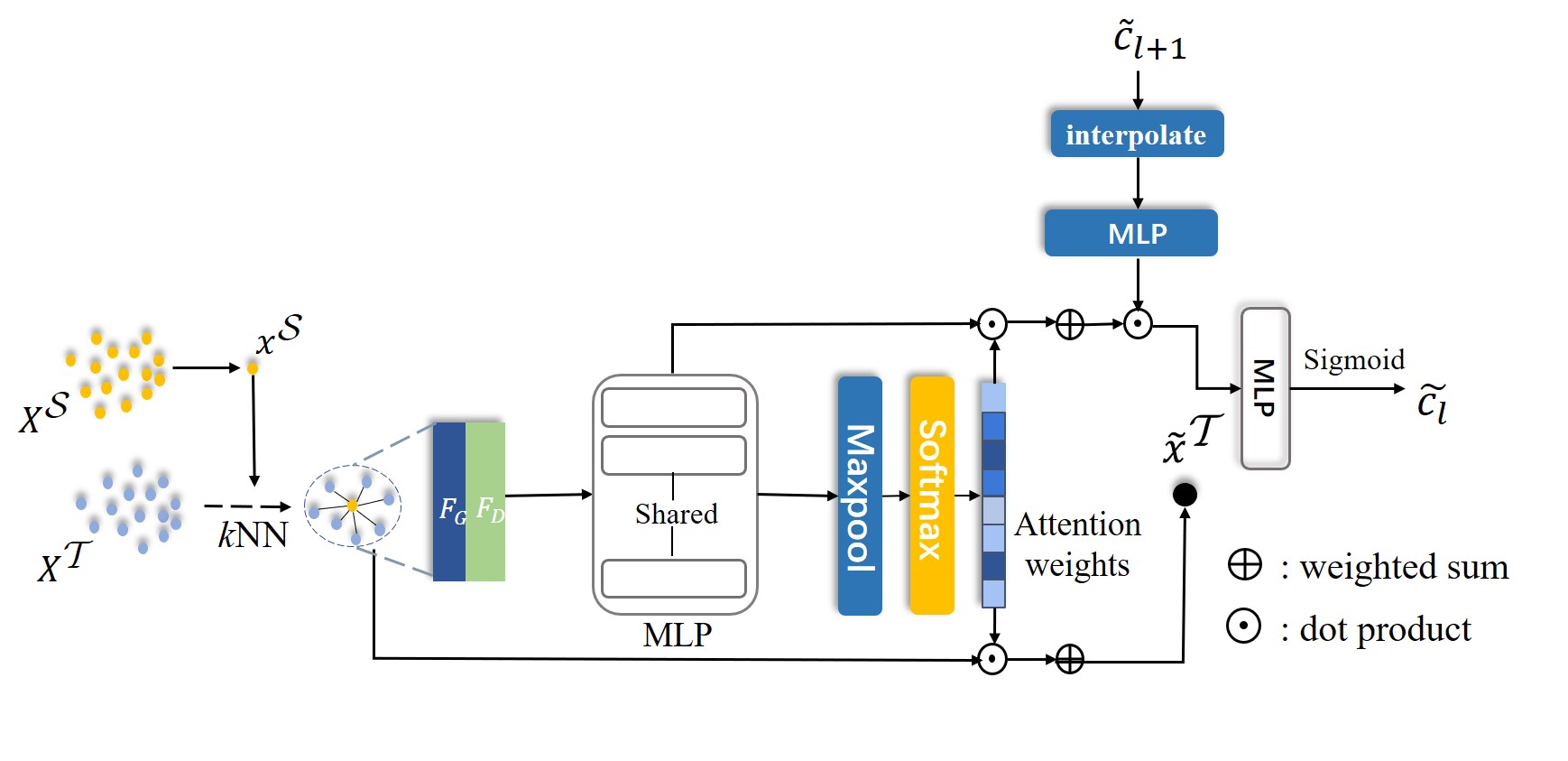}
    \caption{Architecture of correspondence Network in fine registration. Given the source and the target point cloud, along with matching confidence scores of the upper layer, We firstly upsample the confidence scores to obtain an initial mask and utilize the single-soft matching module to generate denser correspondence. The detailed process is described in \cref{3.3}.}
    \label{fig4}
\end{figure}
\subsubsection{Similarity Features}
\label{3.2.2}
\noindent\textbf{Bilateral consensus.} Bilateral consensus is used to describe whether the center point of a cluster and its neighboring points are mutual nearest neighbors during the soft matching process, which has been explained in detail in \cite{rocco2018neighbourhood,lu2021hregnet}. To ensure efficiency, this similarity is only calculated in the second soft matching module.\\
\textbf{Feature consistency similarity.} Feature consistency similarity is computed in both soft matching processes. For a point $x_i^\mathcal{S}$ in the source point cloud $X^\mathcal{S}$ and a cluster $\{x_i^1, x_i^2,...,x_i^k\}$ in the target or updated target point cloud, we denote their descriptors as $d_i^S$ and $d_i^k$, respectively. The cosine similarity between them can be calculated as:
\begin{equation}\label{eq:2}
s_i^k = \frac{{d_i^\mathcal{S} \cdot d_i^k}}{{\| d_i^\mathcal{S} \| \cdot \| d_i^k \|}}
\end{equation}
\hspace*{1em}then the similarity is normalized as:
\begin{equation}\label{eq:3}
s_i^k = \frac{s_i^k}{\max_{m} s_i^m}
\end{equation}
\hspace*{1pc}In the first soft matching process, the similarity quantifies the level of similarity between individual points in $X^\mathcal{S}$ and their corresponding points in $X^\mathcal{T}$, while in the second soft matching process, it measures the level of matching between distinct regions in $X^\mathcal{T}$ and keypoint in $X^\mathcal{S}$.
\subsection{Fine registration with hierarchical mask optimization}
\label{3.3}
Following the coarse registration, we leverage two subsequent layers of fine registration to refine the alignment. In each fine registration layer, we firstly transform the source point cloud using the transformation $\mathbf{R}_{l+1}$, $\mathbf{t}_{l+1}$ obtained from the registration in the deeper layer. A single-soft matching network combined with the weighted kabsch algorithm\cite{kabsch1976solution} is employed to obtain refined transformation $\mathbf{\triangle R}_l$ and $\mathbf{\triangle t}_l$. Hence, the output of this layer is given by $\mathbf{R}_l = \mathbf{\triangle R}_l\mathbf{R}_{l+1}$ and $\mathbf{t}_l = \mathbf{\triangle R}_l\mathbf{t}_{l+1} + \mathbf{\triangle t}_l$.\\
\hspace*{1pc}The structure of the fine registration is illustrated in the \cref{fig4}. We employed the single-soft matching module in the fine registration, as detailed in Section \ref{3.2}. Due to the generation of keypoints in the deeper layer from a region in the shallower layer, the high-confidence keypoints in the denser point cloud of the deeper layer correspond to a high-confidence spatial neighborhood region in the shallower layer. We utilized the confidence scores from the deeper layer to design a hierarchical embedding Mask module. Specifically, We adopt the upsampling method inspired by \cite{qi2017pointnet++} to upsample the confidence scores between layers. For each point $x_{l-1}^i$ in $X_{l-1}$, we perform a \textit{k}NN search in $X_{l}$. Then, we compute the initial weights $c_{l-1}^i$ for $x_{l-1}^i$ of $X_{l-1}$ as:
\begin{equation}\label{eq:4}
c_{l-1}^i = \frac{ \sum_{j=1}^{k} w(i,j) \cdot c_l(i,j)}{ \sum_{j=1}^{k}w(i,j)}\\
\end{equation}
where $c_l(i,j)$ represents the $j$-th nearest neighbor point of $x_i$ in the $l$-th layer. $w(i,j)$ represents the weight of that point, calculated as the reciprocal of the distance between the two points. The weights are then fed into a MLP, where they undergo a mapping to the feature dimension, enabling the application of a mask to each channel of attentive feature map. In addition, considering the effectiveness of global registration in achieving initial alignment of point clouds, all \textit{k}NN searches in the fine registration are performed in Euclidean space rather than in descriptor space.

\begin{figure*}
\begin{minipage}[t]{0.6\textwidth}
  \centering
  \begin{subfigure}[t]{1\textwidth}
    \centering
    \includegraphics[width=\textwidth]{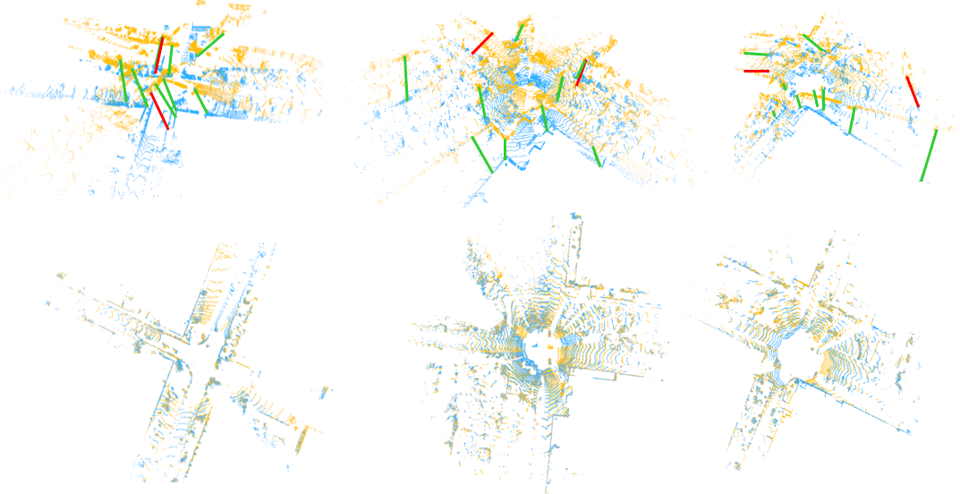}
    \subcaption{Registration result}
  \end{subfigure}
  \label{fig5-a}
\end{minipage}%
\begin{minipage}[t]{0.4\textwidth}
  \centering
  \begin{subfigure}[t]{1\textwidth}
    \centering
    \includegraphics[width=0.98\textwidth]{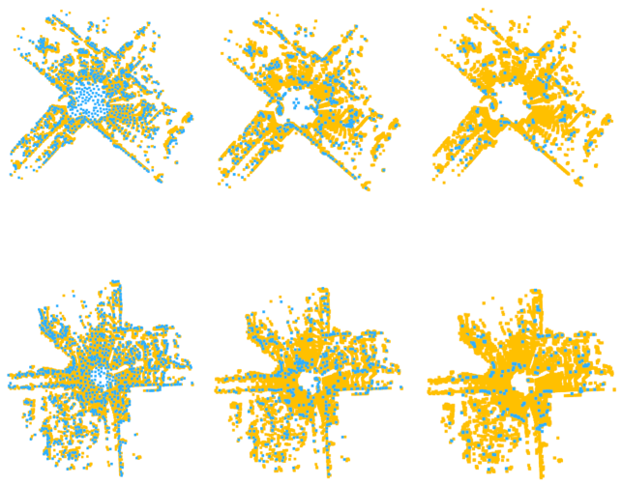}
    \subcaption{Keypoints at each layer}
  \end{subfigure}
  \label{fig5-b}
\end{minipage}
\caption{(a) Qualitative visualization of registration result. We display 3 samples of point cloud registration. The first row display  the top 5 and bottom 5 point correspondences based on their correspondences confidence obtained from the deepest layer's registration. The green lines and the red lines represent inlier and outlier correspondences. The second raw displays the aligned two point clouds. (b) Qualitative visualization of keypoints. We present the keypoints at each layer, with each row representing one point cloud. The yellow points represent the original point cloud, while the blue points indicate the keypoints generated.}
\label{fig5}
\end{figure*}

\subsection{Loss Function}
\label{3.4}
The loss function of HDMNet is composed of translation loss $\mathcal{L}_{trans}$ and rotation loss $\mathcal{L}_{rot}$. The total loss $\mathcal{L}= \mathcal{L}_{trans} + \alpha\mathcal{L}_{rot}$. $\mathcal{L}_{trans}$ and $\mathcal{L}_{rot}$ can be calculated based on the given estimated and ground truth transformations $\mathbf{\hat{R}},\mathbf{\hat{t}}$ and $\mathbf{R},\mathbf{t}$ as
\begin{equation}\label{eq:5}
\mathcal{L}_{trans}=\|\mathbf{t}-\mathbf{\hat{t}}\|_2
\end{equation}
\begin{equation}\label{eq:6}
\mathcal{L}_{rot}=\|\mathbf{\hat{R}^TR}-\mathbf{I}\|_2
\end{equation}\\
where $\mathbf{I}$ denotes identity matrix.
\section{Experiments}
\label{4}
\subsection{Datasets}
\label{4.1}
We conducted experiments on two outdoor LiDAR point cloud datasets: the KITTI odometry dataset\cite{geiger2012we}, the NuScenes dataset\cite{caesar2020nuscenes}. For both datasets, we utilized the point cloud pairs provided in \cite{lu2021hregnet} for training, validation, and testing of the proposed method.
\subsection{Implementation details}
\label{4.2}
\noindent\textbf{Our HDMNet.} In the pre-processing stage, we begin by voxelizing the input point clouds with a voxel size of 0.3m. Subsequently, we randomly sample 16,384 points from the point clouds in the KITTI dataset and 8,192 points in the NuScenes dataset. Our network implementation is based on PyTorch\cite{paszke2019pytorch} and we employed the Adam optimizer\cite{kingma2014adam} to optimize the network parameters. We initialized the learning rate to 0.0095 and implemented a schedule to decrease it by half every 10 epochs. To achieve a balance between rotation and translation, we set the hyperparameter $\alpha$ to 1.8 for the KITTI dataset and 2.1 for the NuScenes dataset. During the network training process, we follow the approach of \cite{lu2020rskdd} to train the detector and descriptor, utilizing the probabilistic chamfer loss as proposed in \cite{li2019usip} and the matching loss as introduced in \cite{lu2020rskdd}. Subsequently, we fix the weights of the keypoint detector and train the entire network by minimizing the rotation and translation errors. The initial layer contains 1024 keypoints, and each subsequent layer has half the number of keypoints compared to the upper layer. The descriptor dimension in the first layer is 64, and in each subsequent layer, it is doubled relative to the upper layer. In both coarse and fine registration stages, the value $k$ of \textit{k}NN search in soft matching module is set to 8.\\
\noindent\textbf{Baseline method.} The performance of the proposed HDMNet is compared with classical methods as well as learning-based methods. All experiments were conducted on an Intel i9-10920X CPU and an NVIDIA RTX 3090 GPU. Following \cite{lu2021hregnet}, we compare the proposed method with 4 representative traditional registration methods: point-to-point and point-to-plane ICP (ICP (P2P) and ICP (P2Pl))\cite{besl1992method}, Fast Global Registration (FGR)\cite{zhou2016fast}, RANSAC\cite{fischler1981random} with FCGF\cite{choy2019fully} as the feature. The maximum iteration number of RANSAC is set to 2×10e6. Additionally, we compare HDMNet with 5 learning-based methods on KITTI dataset and NuScenes dataset, including two object-level registration methods (Deep Closest Point (DCP)\cite{wang2019deep} and IDAM\cite{li2020iterative}), two indoor point cloud registration methods (Feature-metric registration (FMR)\cite{huang2020feature} and Deep Global Registration (DGR)\cite{choy2020deep}), and the state-of-the-art LiDAR point cloud registration network HRegNet\cite{lu2021hregnet}.

\begin{figure*}[t]
\begin{minipage}[t]{1\textwidth}
  \centering
  \begin{subfigure}[t]{1\textwidth}
    \centering
    \includegraphics[width=\textwidth]{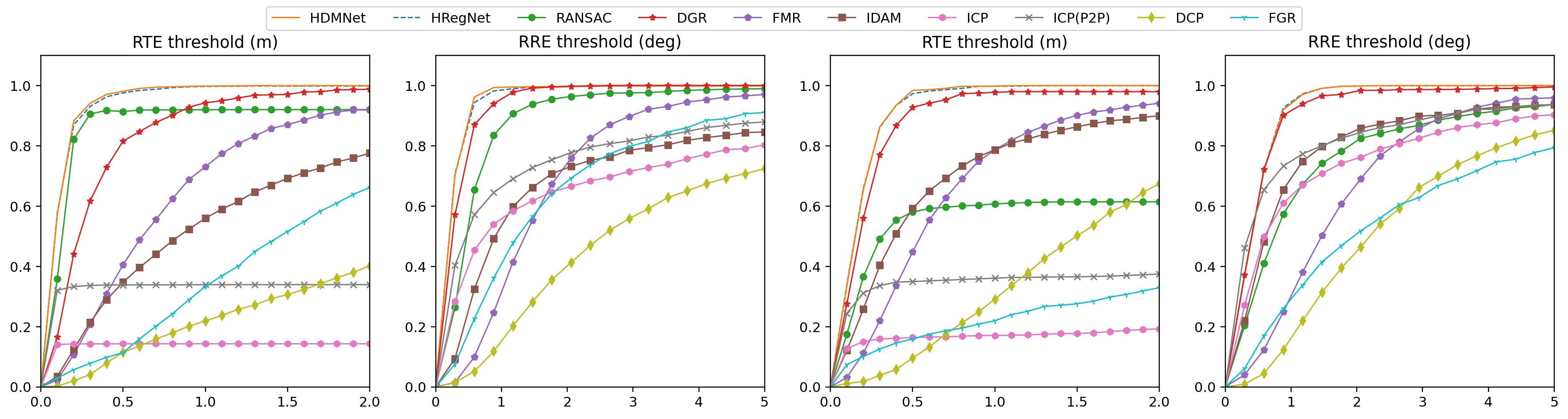}
    \caption{KITTI Dataset\hspace{24em}(b) NuScenes Dataset}
  \end{subfigure}
  \label{fig:main-figure}
\end{minipage}
\label{multi}
\caption{Registration recall with different RRE and RTE thresholds on KITTI dataset and NuScenes dataset.}
\end{figure*}

\begin{table*}
\caption{Registration performance on KITTI dataset and NuScenes dataset.}
\label{tab1}
\fontsize{8}{10}\selectfont
\begin{tabularx}{\linewidth}{@{\extracolsep{\fill}}l|*{4}{X}|*{4}{X}@{}}
\toprule
\multirow{2}{*}{Methods} & \multicolumn{4}{c|}{KITTI dataset} & \multicolumn{4}{c}{NuScenes dataset} \\
\cmidrule{2-5}\cmidrule{6-9}
&RTE(m)&RRE(deg)&Recall&Time(ms)&RTE(m)&RRE(deg)&Recall&Time(ms)\\
\midrule
ICP(P2Point)\cite{besl1992method}&0.045\hspace{0.12em}±\hspace{0.12em}0.054&\underline{0.112\hspace{0.12em}±\hspace{0.12em}0.093}&14.25\%&472.2&0.252\hspace{0.12em}±\hspace{0.12em}0.510&0.253\hspace{0.12em}±\hspace{0.12em}0.502&18.78\%&82.0\\
ICP(P2Plane)\cite{besl1992method}&\underline{0.044\hspace{0.12em}±\hspace{0.12em}0.041}&0.145\hspace{0.12em}±\hspace{0.12em}0.153&33.56\%&461.7&0.153\hspace{0.12em}±\hspace{0.12em}0.296&\underline{0.212\hspace{0.12em}±\hspace{0.12em}0.306}&36.83\%&44.5\\
FGR\cite{zhou2016fast}&0.929\hspace{0.12em}±\hspace{0.12em}0.592&0.963\hspace{0.12em}±\hspace{0.12em}0.807&39.43\%&506.1&0.708\hspace{0.12em}±\hspace{0.12em}0.622&1.007\hspace{0.12em}±\hspace{0.12em}0.924&32.24\%&284.6\\
RANSAC\cite{fischler1981random}&0.161\hspace{0.12em}±\hspace{0.12em}0.093&0.424\hspace{0.12em}±\hspace{0.12em}0.285&\underline{100\%}&459.4&0.171\hspace{0.12em}±\hspace{0.12em}0.128&0.419\hspace{0.12em}±\hspace{0.12em}0.257&99.89\%&187.4\\
\midrule
DCP\cite{wang2019deep}&1.028\hspace{0.12em}±\hspace{0.12em}0.506&2.074\hspace{0.12em}±\hspace{0.12em}1.190&47.29\%&46.4&1.087\hspace{0.12em}±\hspace{0.12em}0.491&2.065\hspace{0.12em}±\hspace{0.12em}1.142&58.58\%&45.5\\
IDAM\cite{li2020iterative}&0.659\hspace{0.12em}±\hspace{0.12em}0.483&1.057\hspace{0.12em}±\hspace{0.12em}0.939&70.92\%&\underline{33.4}&0.467\hspace{0.12em}±\hspace{0.12em}0.410&0.793\hspace{0.12em}±\hspace{0.12em}0.783&87.98\%&\underline{32.6}\\
FMR\cite{huang2020feature}&0.657\hspace{0.12em}±\hspace{0.12em}0.483&1.493\hspace{0.12em}±\hspace{0.12em}0.847&90.58\%&85.5&0.603\hspace{0.12em}±\hspace{0.12em}0.391&1.610\hspace{0.12em}±\hspace{0.12em}0.974&92.06\%&61.1\\
DGR\cite{choy2020deep}&0.322\hspace{0.12em}±\hspace{0.12em}0.320&0.374\hspace{0.12em}±\hspace{0.12em}0.302&98.71\%&1496.6&0.211\hspace{0.12em}±\hspace{0.12em}0.183&0.476\hspace{0.12em}±\hspace{0.12em}0.430&98.41\%&523.0\\
HRegNet\cite{lu2021hregnet}&0.056\hspace{0.12em}±\hspace{0.12em}0.075&0.178\hspace{0.12em}±\hspace{0.12em}0.196&99.77\%&106.2&0.122\hspace{0.12em}±\hspace{0.12em}0.112&0.273\hspace{0.12em}±\hspace{0.12em}0.197&100\%&87.3\\
\midrule
HDMNet&\textbf{0.050\hspace{0.12em}±\hspace{0.12em}0.057}&\textbf{0.159\hspace{0.12em}±\hspace{0.12em}0.152}&\textbf{99.85\%}&\textbf{120.2}&\underline{\textbf{0.114\hspace{0.12em}±\hspace{0.12em}0.102}}&\textbf{0.274\hspace{0.12em}±\hspace{0.12em}0.206}&\underline{\textbf{100\%}}&\textbf{102.9}\\
\bottomrule
\end{tabularx}
\end{table*}


\subsection{Evaluation}
\label{4.3}
\subsubsection{Qualitative visualization}
\label{4.3.1}
We showcase the final registration results in \cref{fig5}(a). In the first row, we select the top 5 and bottom 5 point correspondences based on their correspondences confidence obtained from the deepest layer's registration. If the relative positional error is less than the distance threshold $\epsilon_d =$ 1m, the corresponding keypoints are considered inliers. This classification is visually represented by the green and red lines, respectively denoting the inlier and outlier correspondences. We observed that erroneous correspondences are consistently present among the point correspondences ranked in the bottom 5 in terms of confidence, whereas the top 5 correspondences were generally considered inliers. This indicates that the confidence scores predicted by the network have the potential to reject unreliable correspondences. The second row displays the aligned point clouds.\\
\hspace*{1pc}In addition, in  \cref{fig5}(b), we present the keypoints at each layer. It can be observed that as the layers become deeper, the number of keypoints decreases while their reliability increases. The keypoints in the deepest layer are primarily distributed along the road lines, suggesting their higher reliability. This demonstrates the validity of the hierarchical structure and the effective utilization of robust keypoints in the deepest layers for achieving global registration.
\subsubsection{Quantitative evaluation}
\label{4.3.2}
The evaluation of our approach is conducted based on the relative translation error (RTE) and relative rotation error (RRE), which can be calculated as \cref{eq:5} and $\mathrm{arccos}(\mathrm{Tr}(\mathbf{\hat{R}^TR}-1)/2)$, respectively. where $\mathbf{\hat{R}}^T$ and $\mathbf{R}$ are the estimated and ground truth rotation matrix. In accordance with \cite{lu2021hregnet}, we employ the registration recall to measure the success rate of our registration. For a more detailed comparison of the registration performance, we calculate the average RRE, RTE and
display the results in \cref{tab1}. The average RTE and RRE are only calculated for successful registrations. The values of RTE and RRE are set to $\epsilon_{trans} =$ 2m and $\epsilon_{rot} =$ 5deg, respectively. \\
\hspace*{1pc}The results reveal that without providing a good initial transformation, the ICP algorithm easily falls into local optima, leading to a high failure rate . The recall of FGR is also low, rendering it impractical for applications. The incorporation of the RANSAC algorithm into traditional methods improves both the success rate and accuracy of registration. However, due to multiple iterations, this method's runtime on the KITTI dataset is nearly five times longer than our approach, and its accuracy is also lower than us. Regarding learning-based methods, algorithms designed for indoor or object-level point cloud registration are clearly inadequate for large-scale outdoor scenes. DCP achieves a recall rate of less than 60\% on both the KITTI and NuScenes datasets. IDAM performs better than DCP, but its RTE, RRE are still significantly higher than our algorithm. Moreover, the registration success rates of these algorithms range from 50\% to 70\%, indicating poor feature representation capabilities and limited adaptability to large-scale scenes. FMR does not perform fine registration after global alignment, resulting in lower success rate and accuracy, despite its shorter computation time. DGR has the highest computational cost, but still falls short of achieving sufficient accuracy. HRegNet achieves higher accuracy than other algorithms. However, our proposed HDMNet achieves a reduction of over 12.0\% in RTE and 10.7\% in RRE on the KITTI dataset, with a higher registration recall rate. Additionally, the RTE and RRE distributions exhibit a smaller standard deviation compared to other methods.
\begin{figure}[t!]
    \centering
    \includegraphics[width=1\linewidth]{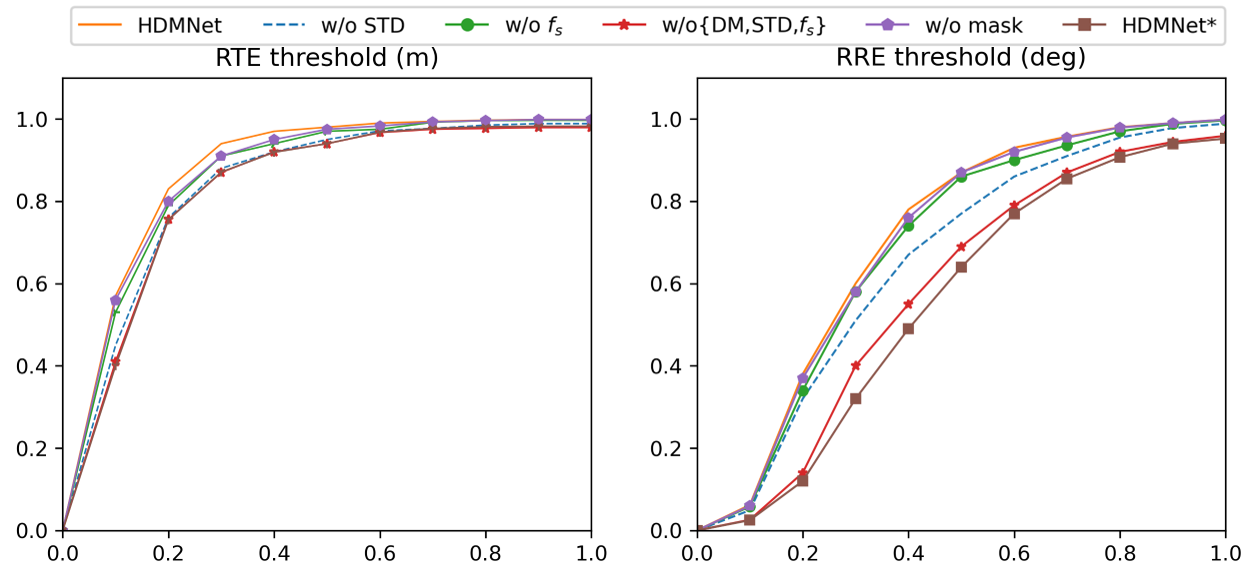}
    \caption{Registration recall of different models on KITTI dataset. STD: Sparse-to-Denser matching strategy. DM: double-soft matching network. $f_s$: feature consistency similarity.}
    \label{ab}
\end{figure}

\begin{table}
\caption{Ablation studies on KITTI dataset.}
\label{tab2}
\centering
\fontsize{8}{10}\selectfont
\begin{tabular}{@{}p{0.22\linewidth} p{0.18\linewidth} p{0.18\linewidth} p{0.08\linewidth} p{0.12\linewidth}@{}}
\toprule
Model&RTE(m)&RRE(deg)&Recall&Time(ms)\\
\midrule
w/o STD&0.064\hspace{0.1em}±\hspace{0.1em}0.118&0.196\hspace{0.1em}±\hspace{0.1em}0.311&98.87\%&110.6\\
w/o $f_s$&0.053\hspace{0.1em}±\hspace{0.1em}0.086&0.165\hspace{0.1em}±\hspace{0.1em}0.191&99.68\%&118.1\\
w/o\{DM,STD,$f_s$\}&0.071\hspace{0.1em}±\hspace{0.1em}0.132&0.192\hspace{0.1em}±\hspace{0.1em}0.272&97.92\%&91.8\\
\midrule
w/o mask&0.053\hspace{0.1em}±\hspace{0.1em}0.068&0.163\hspace{0.1em}±\hspace{0.1em}0.172&99.85\%&117.0\\
\midrule
HDMNet&\textbf{0.050\hspace{0.1em}±\hspace{0.1em}0.057}&\textbf{0.159\hspace{0.1em}±\hspace{0.1em}0.152}&\textbf{99.85\%}&\textbf{120.2}\\
HDMNet*&0.076\hspace{0.1em}±\hspace{0.1em}0.146&0.195\hspace{0.1em}±\hspace{0.1em}0.297&98.24\%&90.1\\
\bottomrule
\end{tabular}
\end{table}
\subsection{Ablation study}
\label{4.4}
To analyze the impact of our proposed double-soft matching module, similarity feature, and mask-prediction module on the performance, we conducted extensive ablation studies on the KITTI dataset provided in \cite{lu2021hregnet}. The registration recall with different modules is displayed in \cref{ab} and the detailed average RTE and RRE is shown in \cref{tab2} and the calculation settings are the same as that in \cref{tab1}. HDMNet* only utilized a single soft-matching without all the aforementioned modules.\\
\textbf{Feature consistency enhanced double-soft matching network.} The Feature Consistency Enhanced Double-Soft Matching Network module(DM) incorporates the paradigm of double-soft matching and introduces feature consistency similarity and sparse-to-denser(STD) strategy. To analyze their individual contributions, we conduct ablation studies by separately dropping these three modules and retraining the model. Specifically, in the first experiment, we perform double-soft matching without sparse-to-denser(STD) strategy and in the second experiment we exclude the feature consistency similarity. Additionally, we conducted another experiment using only single-soft matching (without all the aforementioned components).\\
\hspace*{1pc}The results demonstrate that the Feature Consistency Enhanced Double-Soft Matching Network significantly reduces the RRE, RTE and the standard deviation of the error distribution. This indicates that the estimated values become increasingly stable. The registration recall without the similarity and STD strategy is noticeably inferior to the full model, highlighting the importance of these two strategies. Furthermore, the inclusion of STD strategy further improves the performance, indicating that relying solely on double-soft matching may result in the neglect of the distinctive features of the original target keypoints.\\
\textbf{Mask-prediction}. We drop the Mask Prediction module and retrain the network for comparison with the full model.
The experimental results demonstrate the effectiveness of this approach. Specifically, this strategy leads to a reduction in both RRE and RTE while introduces minimal additional parameters and computations, resulting in a nearly unchanged running time. These findings validate the efficacy of this module, as it contributes to the improvement of accuracy without compromising computational efficiency.
\section{Conclusion}
In this paper, we propose a hierarchical network for outdoor LiDAR point cloud registration, composed of global local-to-local registration and efficient fine registration. To establish reliable correspondences between keypoints, we introduce a double-soft matching network and incorporate feature consistency similarity in the matching process. Additionally, we utilize the confidences of correspondences from deep layer to mask keypoints in the corresponding regions of shallow layers. Abundant ablation experiments demonstrate the effectiveness of our feature consistency enhanced double-soft matching network and mask-prediction module. Furthermore, through extensive experiments on two large-scale outdoor LiDAR point cloud datasets, we have achieved remarkable levels of precision and efficiency with the proposed HDMNet.\\
\\
\textbf{Acknowledgments:}\hspace*{0.3em}This work is supported by the National Natural Science Foundation of China (No.62372329), in part by Shanghai Rising Star Program (No.21QC1400900), Tongji-Qomolo Autonomous Driving Commercial Vehicle Joint Lab Project and Xiaomi Young Talents Program. 

\newpage
{\small
\bibliographystyle{ieee_fullname}
\bibliography{hdmnet}
}

\end{document}